\documentclass[lettersize,journal,x11names]{IEEEtran}
\usepackage{amsmath,amsfonts}
\usepackage{algorithmic}
\usepackage{algorithm}
\usepackage{array}
\usepackage[caption=false,font=footnotesize,labelfont=rm,textfont=rm]{subfig}
\usepackage{textcomp}
\usepackage{stfloats}
\usepackage{url}
\usepackage{verbatim}
\usepackage{graphicx}
\usepackage{cite}
\usepackage{multirow}
\usepackage[dvipsnames]{xcolor}
\usepackage{amsmath,amssymb,bbm,xcolor}
\usepackage[numbers]{natbib}
\usepackage{booktabs}
\usepackage{newfloat}
\usepackage{listings}
\usepackage{threeparttable}
\usepackage{arydshln}
\usepackage{siunitx}
\DeclareCaptionStyle{ruled}{labelfont=normalfont,labelsep=colon,strut=off}

\newcommand{\ignore}[1]{}

\usepackage[normalem]{ulem}

\begin{document}

\title{Speech Translation Refinement using Large Language Models}

\author{Huaixia Dou, Xinyu Tian, Xinglin Lyu, Jie Zhu, Junhui Li, Lifan Guo
\thanks{Huaixia Dou, Xinyu Tian, Xinglin Lyu, and Junhui Li are with the School of Computer Science and Technology, Soochow University, Suzhou, China (e-mail: \{20225227069, xytian, xllv2020\}@stu.suda.edu.cn; lijunhui@suda.edu.cn)}
\thanks{Jie Zhu and Lifan Guo are with the Alibaba Cloud Computing, Hangzhou, China (e-mail: zhujie951121@gmail.com; lifan.lg@alibaba-inc.com)}
}

\markboth{Journal of \LaTeX\ Class Files,~Vol.~14, No.~8, August~2021}%
{Shell \MakeLowercase{\textit{et al.}}: A Sample Article Using IEEEtran.cls for IEEE Journals}


\maketitle
\markboth{}{}

\begin{abstract}
Recent advancements in large language models (LLMs) have demonstrated their remarkable capabilities across various language tasks. Inspired by the success of text-to-text translation refinement, this paper investigates how LLMs can improve the performance of speech translation by introducing a joint refinement process. Through the joint refinement of speech translation (ST) and automatic speech recognition (ASR) transcription via LLMs, the performance of the ST model is significantly improved in both training-free in-context learning and parameter-efficient fine-tuning scenarios. Additionally, we explore the effect of document-level context on refinement under the context-aware fine-tuning scenario. Experimental results on the MuST-C and CoVoST 2 datasets, which include seven translation tasks, demonstrate the effectiveness of the proposed approach using several popular LLMs including GPT-3.5-turbo, LLaMA3-8B, and Mistral-12B. Further analysis further suggests that jointly refining both transcription and translation yields better performance compared to refining translation alone. Meanwhile, incorporating document-level context significantly enhances refinement performance. We release our code and datasets on GitHub\footnote{\url{https://github.com/world1tree/SpeechTranslationRefinement}}.
\end{abstract}


\section{Introduction}
\IEEEPARstart{S}{peech}-to-text translation (ST) refers to the process of converting the audio of a source language into written text in a target language. Despite impressive progress in ST, the performance of ST models based on either the cascade \cite{zhang_etal_acl_2019_lattice,sperber_etal_acl_2020_speech,lam_etal_icassp_2021_cascaded} or end-to-end ~\cite{ye_etal_interspeech_2021_xstnet,fang_etal_acl_2022_stemm,lei_etal_acl_2023_ckdst} frameworks still significantly lags behind that of text-to-text translation models, leaving much room for improvement. Meanwhile, recent studies in text-to-text translation have shown that applying post-editing refinement via large language models (LLMs) can greatly improve the fluency and naturalness of the final translation. For example,~\citet{chen-etal-2024-iterative} propose iterative prompts to enable an LLM to self-correct translations while~\citet{raunak_etal_emnlp_2023_leveraging} introduce the intermediate reasoning chain before generating the refinement.
Inspired by these advancements, this paper explores enhancing the performance of an ST model by introducing LLM-based refinement.

\begin{figure}[t]
\centering
\includegraphics[width=\columnwidth, trim={0cm 0cm 0cm 0cm}, clip]{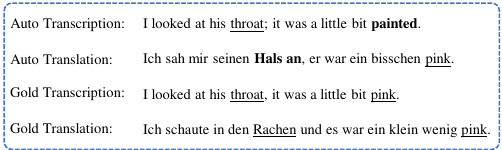}
\caption{Illustration of automatic transcription and translation from ASR and ST models. In this example, errors (\textbf{Bold} text) in both the automatic transcription and translation can be potentially corrected mutually.}
\label{fig:example}
\end{figure}

Unlike studies in text-to-text translation refinement that take clean source text as input, this paper addresses the challenge of refining speech translation where the source text, produced by automatic speech recognition (ASR), contains inevitable errors. Fortunately, even with these errors, the transcription (i.e., ASR output) and the translation (i.e., ST output) can often help correct each other. As shown in Figure~\ref{fig:example}, the error~\textit{Hals an} in the automatic translation can be corrected by consulting the ASR transcription, which accurately recognizes it as~\textit{throat}. Therefore, we propose a joint refinement approach to enhance ST model performance by refining both ASR and ST outputs simultaneously.

Furthermore, this paper proposes applying LLM-based ST refinement in three scenarios: in-context learning, context-agnostic fine-tuning, and context-aware fine-tuning.\footnote{In this paper, \textit{context-agnostic} refers to not incorporating document-level (i.e., inter-sentence) context, while \textit{context-aware} refers to modeling document-level context.} Regarding in-context learning, a dedicated prompt design featuring task descriptions with examples that demonstrate the task. For context-agnostic fine-tuning, LLMs used for refinement can be fine-tuned using parameter-efficient fine-tuning methods like LoRA~\cite{hu_etal_iclr_2022_lora}. In context-aware fine-tuning, we aim to improve the LLMs' robustness to ASR and ST errors by incorporating document-level context. To evaluate the effectiveness of the proposed approach, we conduct extensive experiments on the MuST-C and CoVoST 2 datasets. Several representative LLMs are examined, including GPT-3.5-turbo~\cite{brown_etal_gpt3.5_2020_nips} for the in-context learning scenario, as well as LLaMA3-8B~\cite{dubey_etal_arxiv_2024_llama3} and Mistral-12B~\cite{online_2024_mistral} for both fine-tuning scenarios. Experimental results demonstrate that refining both translation and transcription together achieves better performance compared to refining translation alone. Additionally, incorporating document-level context into refinement can further improve performance. For instance, under the context-aware fine-tuning scenario, Mistral-12B achieves an absolute improvement of 2.98 to 4.22 in BLEU and 0.0450 to 0.0625 in COMET on the MuST-C dataset. 

Overall, the main contributions of this paper can be summarized as follows:   
\begin{itemize}
\item We propose a novel approach to enhance ST model performance by introducing joint LLM-based refinement for both ASR and ST outputs. To the best of our knowledge, this is the first exploration in the ST research field.\footnote{\citet{koneru-etal-iwslt-2024-blending} also explore speech translation refinement, but differs in that they focus on refining ASR and ST separately, whereas our approach jointly refines both tasks. Additionally, we explore both in-context learning and fine-tuning scenarios, extending the analysis to include document-level context.}

\item We adopt the proposed LLM-based refinement across several scenarios, including in-context learning, context-agnostic fine-tuning, and context-aware fine-tuning, to improve the performance of state-of-the-art ST models.

\item We verify the effectiveness of the proposed approach across a wide range of ST tasks, including several translation directions, using several popular LLMs such as GPT-3.5-turbo, LLaMA3-8B and Mistral-12B. 
\end{itemize}

\section{Related work}
\noindent\textbf{Speech-to-Text Translation.} Traditional speech-to-text Translation (ST) systems are typically implemented via a cascade of speech recognition and text translation stages~\cite{zhang_etal_acl_2019_lattice,sperber_etal_acl_2020_speech,lam_etal_icassp_2021_cascaded}, which might incur high latency and error propagation. By contrast, the end-to-end framework has received much attention as it requires no intermediate steps. Existing studies in this line usually utilize the strengths of text translation to enhance ST within a multi-task framework~\cite{ye_etal_interspeech_2021_xstnet}. Techniques such as pre-training~\cite{wang_etal_acl_2020_curriculum,alinejad_sarkar_acl_2020_effectively,tang_etal_acl_2022_unified,zhang_etal_emnlp_2022_speechut}, data augmentation~\cite{pino_etal_iwslt_2019_harnessing,pino_etal_interspeech_2020_self_training,lam_etal_acl_2022_sample}, contrastive learning~\cite{ye_etal_naacl_2022_const,zhang_etal_emnlp_2022_fcgcl,ouyang_etal_acl_2023_waco}, sequence mixup~\cite{fang_etal_acl_2022_stemm,yin_etal_emnlp_2023_improving,zhang_etal_acl_2023_simple,zhou_etal_acl_2023_cmot}, knowledge distillation~\cite{tang_etal_acl_2021_improving,lei_etal_acl_2023_ckdst}, and regularization~\cite{han_etal_acl_2023_modality,gao_etal_naacl_2024_empirical} are widely employed to boost the performance of ST systems. With the rise of LLMs, recent studies have explored combining speech foundation models with LLMs for speech translation~\cite{wu_etal_asru_2023_decoder,chen_etal_icassp_2024_salm,tang_etal_iclr_2024_salmonn,Qwen-Audio,Qwen2-Audio,fathullah-etal-naacl-2024-audiochatllama}. \citet{gaido-etal-acl-2024-st-with-sfm} outlines the typical process, which consists of five components: a speech foundation model for extracting high-level speech representations, a length adapter for compressing speech features, a modality adapter for embedding space mapping, a prompt-speech merger for combining speech, and text prompts and a LLM for generating translation.~\citet{hu-etal-acl-2024-gentranslate} proposes a new generative paradigm for translation tasks, which harnesses the rich information embedded in diverse N-best hypotheses through the use of LLMs to generate higher-quality translation results.

\begin{figure*}[t]
\centering
\includegraphics[width=0.8\textwidth, trim={0cm 0cm 0cm 0cm}, clip]{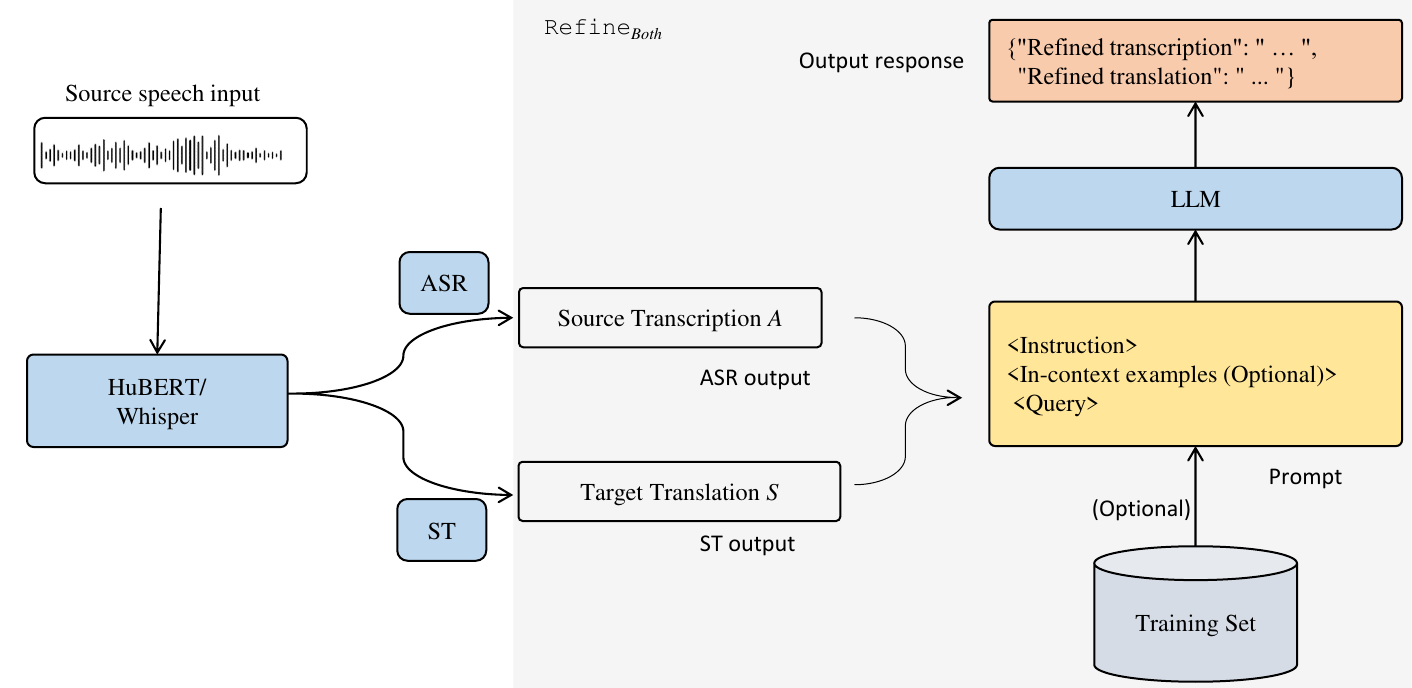}
\caption{Pipeline for the joint refinement. \texttt{Refine}$_{Both}$ is on the right part of the pipeline (highlighted in gray).}
\label{fig:illustration}
\end{figure*}

\noindent\textbf{Text-to-Text Translation Refinement.}
Translation refinement (post-editing) aims to edit the output of MT to produce better translation results. Due to data scarcity, traditional automatic post-editing (APE) relies on manual annotations or large amounts of synthetic data~\cite{chatterjee_etal_wmt_2018_ape,zhang_etal_ACL_2023_lexical}. Recent research indicates that prompting LLMs for APE — particularly prompt-based approaches — can reduce the need for large volumes of training data. ~\citet{chen-etal-2024-iterative} refine translations iteratively with GPT-3.5, which reduces BLEU and chrF++ scores but result in similar or higher COMET scores. \citet{raunak_etal_emnlp_2023_leveraging} refine translations in both Chain of Thought (COT) and non-COT scenarios. \citet{feng_etal_arxiv_2024_improving} propose an LLM-based self-refinement translation framework to improve translation quality across a wide range of languages. To achieve better APE performance, some studies also fine-tune LLMs using supervised data. ~\citet{ki_carpuat_naacl_2024_guiding} leverage LLMs to automatically post-edit translation using\ignore{ multidimensional quality metrics (MQM)} feedbacks\ignore{, refining the output through different prompting strategies and fine-tuning with LLaMA2~\cite{touvron_etal_arxiv_2023_llama2}}. ~\citet{koneru_etal_naacl_2024_contextual} show that fine-tuning LLMs for APE can lead to significant improvements in both sentence and document-level metrics while also generalizing well to out-of-domain data. Inspired by success in text-to-text translation refinement, this paper explores speech translation refinement, a topic that has not been explored before. Specifically, we propose a joint refinement approach that simultaneously addresses errors in both translation and transcription outputs. Similar to the aforementioned translation refinement methods, our approach introduces some latency. However, it is important to emphasize that this latency is accompanied by significant improvements in speech translation quality after refinement.

\section{Approach}
We first describe the task of joint refinement for translation and transcription in Section~\ref{sec:task}. Then we detail our approach to joint refinement in the in-context learning scenario in Section~\ref{sec:in_context}, followed by our approaches to joint refinement in both fine-tuning scenarios in Section~\ref{sec:fine_tuning}.

\subsection{Joint Refinement for Translation and Transcription} 
\label{sec:task} 

We introduce the task of joint refinement for translation and transcription, denoted as \texttt{Refine}$_{Both}$. This task aims to refine both a source-side transcription $A$ (i.e., ASR output) and a target-side translation $S$ (i.e., ST output) by generating an improved transcription $A'$ and an improved translation $S'$, i.e., $\left(A, S\right)\rightarrow \left(A', S'\right)$. Note that during the generation, $A'$ is predicted first, followed by $S'$. This allows the improved transcription $A'$ to be used for generating $S'$.

The detailed pipeline for joint refinement with LLMs is shown in Figure~\ref{fig:illustration}. It starts with the source input speech, which is encoded using a pre-trained speech recognition model such as HuBERT~\cite{hsu_etal_taslp_2021_hubert} or Whisper~\cite{radford_etal_icml_2023_whisple}. The ASR model then generates a transcription $A$ while the ST model produces a translation $S$.\ignore{ In contrast to previous research on translation refinement that uses noise-free source-side sentences, our approach operates with noisy ASR transcriptions.} In the \texttt{Refine}$_{Both}$ task, we construct a prompt using the transcription $A$, the translation $S$, and the retrieved examples. We then request the LLM to generate an output that includes both a refined transcription and a refined translation.

\paragraph{Two Contrasting Tasks} To better illustrate the effects of joint refinement, we compare \texttt{Refine}$_{Both}$ with two contrasting tasks:

\begin{itemize}
\item \texttt{Refine}$_{ST}$: This task focuses solely on refining the translation by generating an improved translation $S'$, i.e., $\left(A, S\right) \rightarrow \left(S'\right)$. It improves the translation without taking into account potential errors in the transcription $A$. 
\item \texttt{Paraphrase}$_{ST}$: This task refines the translation without using the ASR transcription, i.e., $\left(S\right) \rightarrow \left(S'\right)$. Similar to \citet{chen-etal-2024-iterative}, this task serves as a contrasting experiment to translation prompting.
\end{itemize}

In the following sections, we use \texttt{Refine}$_{Both}$ to demonstrate the proposed in-context learning and task-specific fine-tuning processes. It is important to note that similar in-context learning and fine-tuning approaches can be applied to \texttt{Refine}$_{ST}$ and \texttt{Paraphrase}$_{ST}$ as well.

\begin{figure*}[t]
\centering
\includegraphics[width=\textwidth, trim={0cm 0cm 0cm 0cm}, clip]{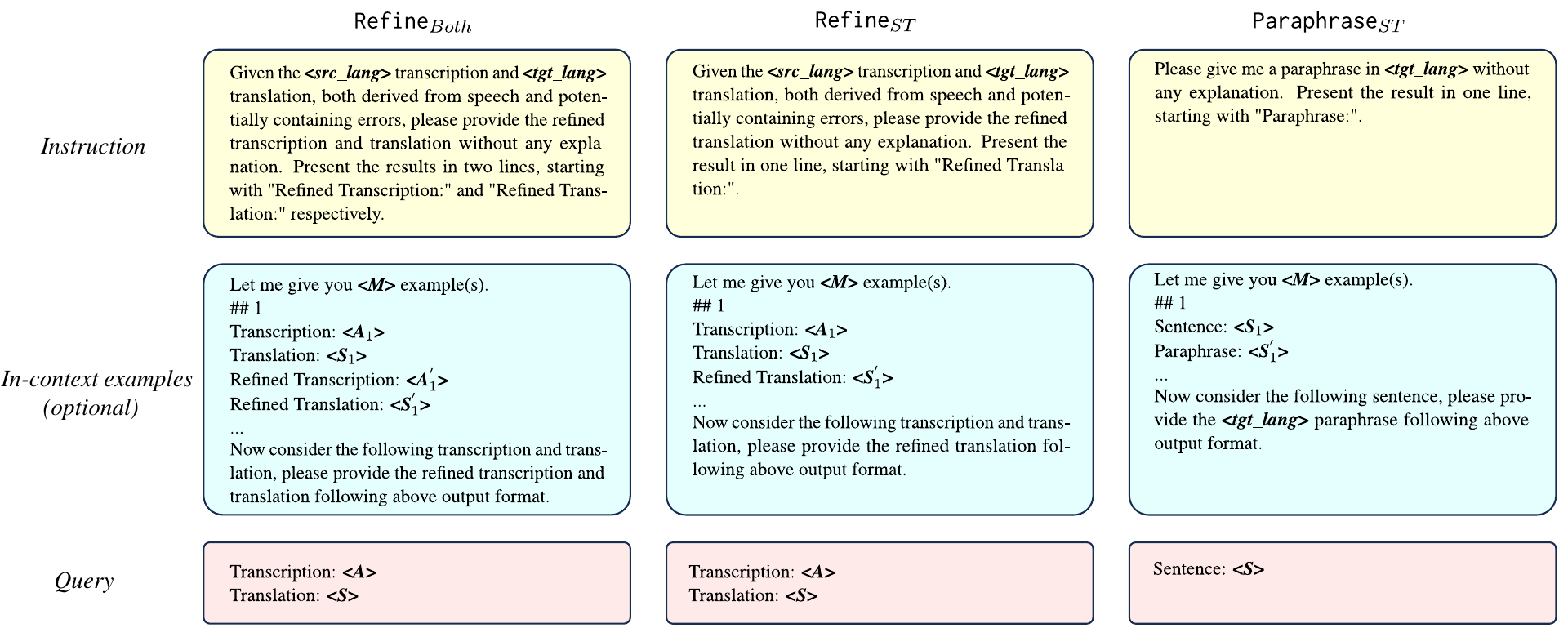}
\caption{Prompts for ST refinement using LLMs, including instruction, optional in-context examples (used only for in-context learning), and query. The placeholders ``\textbf{\textit{\textless var\textgreater}}'' are replaced with their corresponding content. See Appendix~\ref{apdx:prompt_example} for detailed examples.}
\label{fig:prompt1}
\end{figure*}

\subsection{Joint Refining via In-Context Learning}
\label{sec:in_context}

As shown in Figure~\ref{fig:prompt1}, a prompt consists of three main parts: \textit{instruction}, \textit{in-context examples}, and \textit{query}. 

\textit{Instruction} provides a brief definition of joint refinement and specifies constraints on the output format. \textit{In-context examples} provide additional context to enhance performance. Each example includes an automatic transcription and translation, along with their corresponding refined versions. While these examples provide guidance on what to generate, the number of examples that can be included is constrained by the context length. To select $M$ demonstration examples from the training set, we compare two strategies: one selects examples randomly, while the other selects examples based on the L2 distance between a candidate pair $\left(\hat{A}, \hat{S}\right)$ and the input pair $\left(A, S\right)$. Finally, the \textit{query} prompts the LLMs with the question by specifying the automatic transcription and translation as input parameters. The basic form of the prompt is the concatenation of an \textit{instruction}, optional \textit{in-context examples}, and a \textit{query}.

\subsection{Joint Refining via Fine-Tuning}
\label{sec:fine_tuning}

We first present our context-agnostic fine-tuning, which assumes that both the transcription $A$ and the translation $S$ are at the sentence level (Section~\ref{sec:agnostic_fine_tuning}). Then we extend it to context-aware fine-tuning by expanding the scope from a single sentence to multiple sentences (Section~\ref{sec:aware_fine_tuning}).

\subsubsection{Context-Agnostic Fine-Tuning}
\label{sec:agnostic_fine_tuning}

Fine-tuning is effective when only a small amount of training data is available for supervised LLM training. To optimize the refinement capabilities of LLMs, we adopt a two-stage strategy.

\begin{figure}[t]
\centering
\includegraphics[width=0.40\textwidth, trim={0cm 0cm 0cm 0cm}, clip]{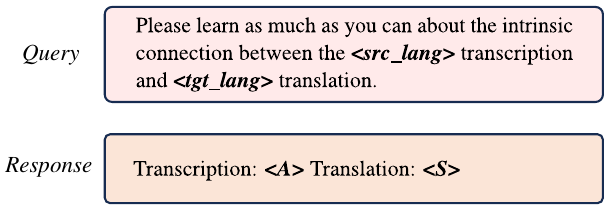}
\caption{Prompt used in the first stage of fine-tuning.}
\label{fig:prompt2}
\end{figure}

\begin{itemize}
    \item Stage 1: Inspired by~\citet{gururangan_etal_acl_2020_dont}, we establish the intrinsic connection between the source-side transcription and target-side translation by fine-tuning LLMs to generate both the source-side transcription $A$ and the target-side translation $S$. Figure~\ref{fig:prompt2} demonstrates the prompt and its output response for this stage. 
    \item Stage 2: In this stage, we continue to fine-tune LLMs using the prompt shown in Figure~\ref{fig:prompt1}, with the output response including both refined transcription and translation. Note that no in-context examples are used during the fine-tuning process.
\end{itemize}

\subsubsection{Context-Aware Fine-Tuning}
\label{sec:aware_fine_tuning}

While document-level context has demonstrated benefits for both textual and speech translation~\cite{tiedemann_scherrer_discmt_2017_neural,li-etal-2023-ptransformer,Zhang_etal_acl_2021_beyond}, its effectiveness in translation refinement has not been well explored. To address this gap, we examine the use of document-level context through a simple concatenation-based strategy, which expands the scope of the text to be refined from single sentences to multiple sentences. Specifically, in this setting, both the source-side transcription $A$ and target-side translation $S$ consist of $K$ neighboring sentences, as do their refined counterparts $A'$ and $S'$. The prompts and output responses during the two-stage fine-tuning process remain consistent with those used in context-agnostic fine-tuning. 

During the inference phase, we use chunk-based decoding (CBD)~\cite{Zhang_etal_acl_2021_beyond}, which splits all sentences in a document into non-overlapping chunks, with each chunk concatenating $K$ neighboring sentences. CBD is efficient because it encodes and decodes each sentence only once. To address potential misalignment issues, where the number of output sentences may differ from the number of input sentences, we prepend an index to each sentence, starting from 1 (e.g., \textit{\#1 ... \#2 ... \#3 ...} when $K$ is set to 3).\footnote{In rare cases of misaligned refinement, we use the original input sentences as the refined result.}

\section{Experimentation}

\begin{table}[t]
\caption{Statistics of all datasets in our experiments.}
 \begin{center}
\begin{tabular}{l|ccc|ccc}
\hline
\multirow{2}{*}{\textbf{En}$\rightarrow$} & \multicolumn{3}{c|}{\textbf{\#Sentences}} & \multicolumn{3}{c}{\textbf{\#Documents}} \\
 & \bf Training & \bf Valid & \bf Test & \bf Training & \bf Valid & \bf Test \\
\hline
\multicolumn{7}{c}{\bf MuST-C}\\
\hline
\textbf{De} &  \num{225271} & \num{1418} & \num{2641} & \num{2041} & \num{11} & \num{27} \\
\textbf{Fr} & \num{269248} & \num{1408} & \num{2632} & \num{2458} & \num{11} & \num{27} \\
\textbf{Es} & \num{260041} & \num{1312} & \num{2502} & \num{2512} & \num{11} & \num{27} \\
\hline
\multicolumn{7}{c}{\textbf{CoVoST 2}}\\
\hline
\textbf{De} & \num{289408} & \num{15520} & \num{15518} & - & - & - \\
\textbf{Ca} & \num{289408} & \num{15520} & \num{15519} & - & - & - \\
\textbf{Ar} & \num{289408} & \num{15520} & \num{15519} & - & - & - \\
\textbf{Tr} & \num{289408} & \num{15520} & \num{15519} & - & - & - \\
\hline
\end{tabular}
\end{center}
\label{tbl:dataset}
\end{table}

\label{sec:experiment}
\begin{table*}[!t]
\caption{Experimental results on MuST-C dataset for translation refinement (measured by BLEU and COMET) and transcription refinement (measured by WER).}
\begin{center}
\resizebox{\linewidth}{!}{
\begin{tabular}{l|l|lll|lll|lll}
\hline
\multirow{2}{*}{\textbf{Task}} & \multirow{2}{*}{\textbf{Model}} & \multicolumn{3}{c|}{\textbf{En$\rightarrow$De}} & \multicolumn{3}{c|}{\textbf{En$\rightarrow$Fr}} & \multicolumn{3}{c}{\textbf{En$\rightarrow$Es}} \\
& & \textbf{BLEU$\uparrow$} & \textbf{COMET$\uparrow$} & \textbf{WER$\downarrow$} & \textbf{BLEU$\uparrow$} & \textbf{COMET$\uparrow$} & \textbf{WER$\downarrow$} & \textbf{BLEU$\uparrow$} & \textbf{COMET$\uparrow$} & \textbf{WER$\downarrow$} \\
\hline
\multirow{2}{*}{-} & CRESS (F\&F) & 29.40 & - & - & 40.10 & - & - & 33.20 & - & - \\
 & CRESS (Ours) & 29.40 & .7669 & 10.65 & 40.00 & .7814 & 10.23 & 33.10 & .7805 & 10.47 \\
\hline
\multicolumn{11}{c}{\bf In-Context Learning}\\
\hline
\multirow{7}{*}{\texttt{Refine}$_{Both}$} & GPT-3.5 (0) & 30.17 & \bf .8147 & 12.67 & 40.22	& \bf .8171 & 12.42	& 34.12	& \bf .8179 & 13.21 \\

& GPT-3.5 (R1) & \bf 30.52 & .8124 & 10.59 & \bf 40.99 & .8158 & 9.95 & 34.03 & .8138 & 10.77 \\
& GPT-3.5 (R3) & 30.38 & .8118 & 10.31 & 40.59 & .8168 & \bf 9.60 & \bf 34.36 & .8154 & 10.29 \\ 
& GPT-3.5 (R5) & 30.46 & .8134 & 10.29 & 40.67 & .8156 & 9.84 & 34.20 & .8156 & 10.02 \\
& GPT-3.5 (T1) & 30.15 & .8101 & 10.82 & 40.85 & .8137 & 10.01 & 34.02 & .8139 & 10.72 \\
& GPT-3.5 (T3) & 30.09 & .8108 & 10.28 & 40.65 & .8161 & 9.66 & 34.20 & .8148 & 10.02 \\
& GPT-3.5 (T5) & 30.41 & .8119 & \bf 10.26 & 40.48 & .8157 & 9.71 & \bf 34.36 & .8153 & \bf 9.89 \\

\hdashline
\multirow{7}{*}{\texttt{Refine}$_{ST}$} & GPT-3.5 (0) & 27.00 & .8045 & - & 35.53 & .8092 & - &	32.12 & .8131 & -  \\
& GPT-3.5 (R1) & 28.95 & .8107 & - & 38.41 & .8142 & - & 33.89 & .8147 & - \\
& GPT-3.5 (R3) & 28.85 & .8121 & - & 38.20 & .8156 & - & \bf 34.16 & \bf .8184 & - \\
& GPT-3.5 (R5) &  28.98	& \bf .8134 & - & 38.61	& \bf .8171 & - & 34.14 & .8174 & -\\
& GPT-3.5 (T1) &  28.69 & .8088 & - & 38.33	& .8118 & - & 33.68 & .8121 & - \\
& GPT-3.5 (T3) & 28.68 & .8110 & - & 38.12 & .8111 & - & 33.64 & .8147 & -  \\
& GPT-3.5 (T5) & \bf 29.04 & .8099 & - & \bf 38.62 & .8150 & - & 34.04 & .8156 & - \\
\hdashline
\multirow{4}{*}{\texttt{Paraphrase}$_{ST}$} & GPT-3.5 (0) & 12.02	& .7579	& - & 13.91	& .7488 & - & 14.27 & .7585 & - \\
& GPT-3.5 (T1) &  19.38	& .7735	& - & 24.38	& .7714 & - & 22.64 & .7758 & - \\
& GPT-3.5 (T3) & 22.65 & .7821 & - & 29.04 & .7843 & - & 26.37 & .7862 & - \\
& GPT-3.5 (T5) & \bf 24.97 & \bf .7848 & - & \bf 31.78	& \bf .7860 & - & \bf 28.74 & \bf .7918 & - \\
\hline
\multicolumn{11}{c}{\bf Context-Agnostic Fine-Tuning}\\
\hline
\multirow{2}{*}{\texttt{Refine}$_{Both}$} & LLaMA3-8B & 32.04 & .8196 & \bf 9.27 & 42.16	& .8202 & 9.03	& 34.99 & .8142 & 9.37 \\
& Mistral-12B & \bf 32.75 & \bf .8216 & 9.29 & \bf 43.59 & \bf .8258 & \bf 8.96 & \bf 35.79 & \bf .8189 & \bf 9.13 \\
\hdashline
\multirow{2}{*}{\texttt{Refine}$_{ST}$} & LLaMA3-8B & 29.92 & .8189 & - & 41.48 & .8212	& - & \bf 34.59 & .8195 & -\\
& Mistral-12B & \bf 30.25 & \bf .8225 & - & \bf 42.61 & \bf .8246 & - & 34.52	& \bf .8223 & - \\
\hdashline
\multirow{2}{*}{\texttt{Paraphrase}$_{ST}$} & LLaMA3-8B &  28.48 & .7975 & - & 40.87 & .8024 & - & \bf 34.30 & .8010 & - \\
& Mistral-12B & \bf 30.11 & \bf .8059 & - & \bf 41.11 & \bf .8076 & - & 33.74 & \bf .8074 & - \\
\hline
\multicolumn{11}{c}{\bf Context-Aware Fine-Tuning ($K=3$)}\\
\hline
\multirow{2}{*}{\texttt{Refine}$_{Both}$} & LLaMA3-8B & 32.43 & .8240 & 8.95 & 42.93 & .8263 & 8.79 & 35.62 & .8241 & 9.20 \\
& Mistral-12B & \bf 33.39 & \bf .8294 & \bf 8.91 & \bf 44.22 & \bf .8293 & \bf 8.43 & \bf 36.08 & \bf .8255 & \bf 8.88\\
\hdashline
\multirow{2}{*}{\texttt{Refine}$_{ST}$} & LLaMA3-8B & 32.2 & .8217 & - & 42.49 & .8262 & - & 34.18 & .8224 & - \\  
& Mistral-12B & \bf 32.6	& \bf .8288 & - & \bf 43.49 & \bf .8314 & - & \bf 35.71 & \bf .8259 & - \\
\hdashline
\multirow{2}{*}{\texttt{Paraphrase}$_{ST}$} & LLaMA3-8B & 30.29	& .8025	& - & 41.34 & .8070 & - & 34.36 & .8048 & - \\
& Mistral-12B & \bf 31.07 & \bf .8122 & - & \bf 42.30 & \bf .8154 & - & \bf 34.80 & \bf .8112 & - \\
\hline
\end{tabular}
}
\end{center}
\label{tbl:mustc_result}
\begin{tablenotes}
\item \textit{Note:}  Here, GPT-3.5 (0) represents zero-shot in-context learning, while GPT-3.5 (R$i$) and GPT-3.5 (T$i$) denote the use of $i$ random or top closest examples for few-shot in-context learning.
\end{tablenotes}
\end{table*}

\subsection{Experimental Setup}
\label{sec:settings}

\paragraph{Datasets} We build the refinement datasets from MuST-C~\cite{di_gangi_etal_naacl_2019_mustc} and CoVoST 2~\cite{wang_etal_arxiv_2020_covost2}. Specifically, we obtain the corresponding automatic transcription and translation on MuST-C and CoVoST 2 by running ASR and ST models. MuST-C, including English (EN) to \{German (DE), French (FR), Spanish (ES)\} three translation directions, is derived from TED talks and provides document-level annotation. Therefore, it can be used for evaluation under the context-aware fine-tuning scenario. In contrast, CoVoST 2 provides only the sentence-level annotation on four translation tasks, including English (EN) to \{German (DE), Catalan (CA), Arabic (AR), Turkish (TR)\}. See Table~\ref{tbl:dataset} for details of data splitting and dataset statistics.

\paragraph{Models} We build our approach on three popular LLMs: GPT-3.5-turbo, LLaMA3-8B\footnote{\url{https://huggingface.co/meta-llama/Meta-Llama-3-8B-Instruct}}, and Mistral-12B\footnote{\url{https://huggingface.co/mistralai/Mistral-Nemo-Instruct-2407}}. GPT-3.5-turbo is renowned for its state-of-the-art performance in in-context learning without the need for parameter fine-tuning. LLaMA3-8B and Mistral-12B are chosen for their popularity as open-source models that can be customized for local fine-tuning.

We employ the best-performing open-source ST models to generate automatic speech translation. Among them, CRESS~\cite{fang_feng_acl_2023_cress} is the top performer on the MuST-C dataset, while SpeechLM-P~\cite{zhang_etal_acm_2024_speechlm} excels on CoVoST 2. Specifically, we use the CRESS model with ASR capability\footnote{We follow the approach in XSTNeT~\cite{ye_etal_interspeech_2021_xstnet} to train it from scratch.} to generate the automatic translation and transcription on MuST-C. We employ SpeechLM-P to generate automatic speech translation on CoVoST 2. Automatic ASR transcriptions for CoVoST 2 are generated using the \texttt{whisper-large-v3}~\cite{radford_etal_icml_2023_whisple} model. 

\paragraph{Training and Inference} For in-context learning without training, we first use the SentenceTransformers library \footnote{\texttt{paraphrase-multilingual-mpnet-base-v2}: \url{https://www.sbert.net/index.html} } to obtain the sentence embeddings $E_A$ and $E_S$ for transcription $A$ and translation $S$. We than concatenate $[E_A, E_S]$ to retrieve examples from the training set and perform inference by concatenating the prompt with the retrieval examples.

For fine-tuning, we use the LLama-Factory~\cite{zheng_etal_acl_2024_llamafactory} framework, setting the LoRA rank to 8 and the scaling parameter to 16.  All models are fine-tuned on 2 Nvidia Tesla V100 GPUs with a batch size of 2, gradient accumulation over 16 steps, and a learning rate of 1e-4 for 2 epochs. We then use the checkpoint with the best performance on the validation set to run the inference with a beam size of 3.

\paragraph{Metrics} For performance evaluation, we report SacreBLEU\footnote{SacreBLEU signature: nrefs:1 $\mid$ bs:1000 $\mid$ seed:12345 $\mid$ case:mixed $\mid$ eff:no $\mid$ tok:13a $\mid$ smooth:exp $\mid$ version:2.0.0}~\cite{post_wmt_2018_call} and COMET\footnote{Unbabel/wmt22-comet-da}~\cite{bosselut_etal_acl_2019_comet} for ST refinement, and WER\footnote{WER is case-sensitive and includes punctuation removal.} for transcription refinement.

\begin{table*}[!t]
\caption{Experimental results on the CoVoST 2 dataset. }
\begin{center}
\resizebox{\linewidth}{!}{
\begin{tabular}{l|l|lll|lll|lll|lll}
\hline
\multirow{2}{*}{\textbf{Task}} & \multirow{2}{*}{\textbf{Model}} & \multicolumn{3}{c|}{\textbf{En$\rightarrow$De}} & \multicolumn{3}{c|}{\textbf{En$\rightarrow$Ca}} & \multicolumn{3}{c|}{\textbf{En$\rightarrow$Ar}} & \multicolumn{3}{c}{\textbf{En$\rightarrow$Tr}}\\
& & \textbf{BLEU$\uparrow$} & \textbf{COMET$\uparrow$} & \textbf{WER$\downarrow$} & \textbf{BLEU$\uparrow$} & \textbf{COMET$\uparrow$} & \textbf{WER$\downarrow$} & \textbf{BLEU$\uparrow$} & \textbf{COMET$\uparrow$} & \textbf{WER$\downarrow$} & \textbf{BLEU$\uparrow$} & \textbf{COMET$\uparrow$} & \textbf{WER$\downarrow$} \\
\hline
- & SpeechLM-P / Whisper & 27.58 & .7579 & 11.09 & 35.89 & .7751 & 11.09 & 21.72 & .7954 & 11.09 & 19.54 & .8007 & 11.09 \\
\hline
\multicolumn{14}{c}{\bf Context-Agnostic Fine-Tuning}\\
\hline
\multirow{2}{*}{\texttt{Refine}$_{Both}$} & LLaMA3-8B & 31.24 & .8230 & 9.16 & 39.62 & .8163 & 9.07 & 23.62	& .8189	& 9.03 & 22.28 & \bf .8366 & 9.25 \\
& Mistral-12B & \bf 32.39 & \bf .8334 & \bf 9.11 & \bf 39.99 & \bf .8165 & \bf 8.90 & \bf 24.04 & \bf .8239 & \bf 8.93	& \bf 22.35 & .8358 & \bf 9.16 \\
\hdashline
\multirow{2}{*}{\texttt{Refine}$_{ST}$} & LLaMA3-8B & 30.02 & .8021 & - & 38.21 & .8005 & - & 22.82 & .8103 & - & \bf 21.92 & \bf .8305 & - \\
& Mistral-12B & \bf 30.51 & \bf .8108 & - & \bf 39.08 & \bf .8075 & - & \bf 23.48 & \bf .8183 & - & 21.87 & .8286 & - \\
\hdashline
\multirow{2}{*}{\texttt{Paraphrase}$_{ST}$} & LLaMA3-8B & 28.43 & .7754 & - & 36.72 & \bf .7823 & - & 21.95 & .7981 & - & 20.19 & \bf .8067 & - \\
& Mistral-12B & \bf 28.93 & \bf .7859 & - & \bf 36.87 & .7814 & - & \bf 22.10 & \bf .8003 & - & \bf 20.22 & .8062 & - \\
\hline
\end{tabular}
}
\end{center}
\label{tbl:covost2_result}
\begin{tablenotes}
\item \textit{Note: }We present the performance of in-context learning settings in Appendix~\ref{apdx:result_covost2}.
\end{tablenotes}
\end{table*}

\subsection{Experimental Results on MuST-C}
\label{sec:result_mustc}

Table~\ref{tbl:mustc_result} compares the performance on MuST-C dataset. The first two rows show that when preparing automatic transcription and translation, we achieve similar ST performance in BLEU as \citet{fang_etal_acl_2022_stemm}. Next let us focus on the \texttt{Refine}$_{Both}$ task and answer the following Q1 to Q3 questions.  

\noindent\textbf{Q1: Does in-context learning improve performance?} Compared to the input \textit{CRESS (Ours)}, in-context learning with GPT-3.5-turbo improves ST refinement, as evidenced by higher BLEU and COMET scores across all three language pairs. However, zero-shot in-context learning significantly degrades transcription refinement performance, whereas few-shot learning has a more modest effect. Additionally, there is no clear benefit of using more in-context examples or whether retrieval-based example selection surpasses random selection.

\noindent\textbf{Q2: Does context-agnostic fine-tuning improve performance?} Yes, context-agnostic fine-tuning enhances performance for both ST and transcription refinement across all three language pairs. For instance, the fine-tuned Mistral-12B shows an absolute improvement of 3.35, 3.59, and 2.69 in BLEU scores, and 0.0547, 0.0444, and 0.0384 in COMET scores for En$\rightarrow$De, En$\rightarrow$Fr, and En$\rightarrow$Es translation refinement, respectively. Additionally, it achieves improvements of 1.36, 1.27, and 1.34 in WER for transcription refinement. Despite having fewer parameters, the fine-tuned LLaMA3-8B and Mistral-12B models outperform GPT-3.5-turbo. 

\noindent\textbf{Q3: Does context improve performance?} Yes, incorporating document-level context enhances performance for both ST and transcription refinement. For instance, expanding refinement from a single sentence to three sentences (i.e., $K=3$) allows Mistral-12B to achieve additional improvements of 0.64, 0.63, and 0.29 in BLEU scores, and 0.0078, 0.0035, and 0.0066 in COMET scores for En$\rightarrow$De, En$\rightarrow$Fr, and En$\rightarrow$Es translation, respectively.

Then, we evaluate the \texttt{Refine}$_{ST}$ and \texttt{Paraphrase}$_{ST}$ tasks, focusing on questions Q4 and Q5.

\noindent\textbf{Q4: How do \texttt{Refine}$_{ST}$ and \texttt{Paraphrase}$_{ST}$ perform?} Compared to the input \textit{CRESS (Ours)}, \texttt{Refine}$_{ST}$ consistently achieves higher COMET scores across all language pairs and in both in-context learning and fine-tuning settings. However, it negatively affects BLEU scores for En$\rightarrow$De and En$\rightarrow$Fr translations in the in-context learning setting. This pattern, where GPT-3.5-turbo achieves lower BLEU scores but higher COMET scores, aligns with \citet{chen-etal-2024-iterative}. In contrast, \texttt{Paraphrase}$_{ST}$ in the in-context learning setting leads to significant BLEU score declines, indicating the importance of using transcription input to avoid semantic drift. Fortunately, \texttt{Paraphrase}$_{ST}$ performs comparably or even better in both context-agnostic and context-aware fine-tuning.

\noindent\textbf{Q5: Does joint performing transcription help translation refinement?} Comparing \texttt{Refine}$_{Both}$ with \texttt{Refine}$_{ST}$ reveals that joint performing transcription results in higher BLEU scores (with all improvements statistically significant at p\textless 0.01 \cite{koehn_emnlp_2004_statistical}), although the COMET scores remain similar. For instance, when context-aware fine-tuning the Mistral-12B, \texttt{Refine}$_{Both}$ achieves BLEU scores that are 2.50, 0.98, and 1.27 points higher than \texttt{Refine}$_{ST}$ for En$\rightarrow$De, En$\rightarrow$Fr, and En$\rightarrow$Es ST refinement, respectively.

\subsection{Experimental Results on CoVoST 2}
\label{sec:result_covost2}

Table~\ref{tbl:covost2_result} shows the performance results on the CoVoST 2 dataset. It is evident that fine-tuning improves performance for both speech translation (ST) and transcription refinement across all four language pairs. Specifically, for the \texttt{Refine}$_{Both}$ task, fine-tuning Mistral-12B results in absolute BLEU score improvements of 4.81, 4.10, 2.32, and 2.81, and COMET score improvements of 0.0755, 0.0414, 0.0285, and 0.0351 for En$\rightarrow$De, En$\rightarrow$Ca, En$\rightarrow$Ar, and En$\rightarrow$Tr, respectively. Additionally, it clearly shows that \texttt{Refine}$_{Both}$ achieves the best result cross the four language pairs, followed by \texttt{Refine}$_{ST}$ and then \texttt{Paraphrase}$_{ST}$.

\section{Analysis}

\label{sec:disc}
In this section, we use MuST-C En$\rightarrow$De as a representative example to demonstrate the effectiveness of our approach, unless stated otherwise. Specifically, Section~\ref{sec:ana:semantic_fluency},~\ref{sec:ana:gpt_eval}, and~\ref{sec:ana:case_study} show that our approach outperforms others from multiple perspectives. Section~\ref{sec:ana:asr_quality},~\ref{sec:ana:two_stage}, and~\ref{sec:ana:result_other_system} compare the performance across different settings.  Finally, Section~\ref{sec:ana:context_len} and~\ref{sec:ana:context_sen} provide further insights into the importance of document-level context for context-aware fine-tuning. 

\subsection{Improvements in Semantics and Fluency}
\label{sec:ana:semantic_fluency}

\begin{table}[!t]
\caption{Semantic and Fluency evaluation results on the MuST-C En$\rightarrow$De test set.}
\begin{center}
\resizebox{\linewidth}{!}{
\begin{tabular}{l|lll|lll}
\hline
\multirow{2}{*}{} & \multicolumn{3}{c|}{\textbf{Transcription}} & \multicolumn{3}{c}{\textbf{Translation}} \\
\textbf{Refinement} & \textbf{BERT-S$\uparrow$} &\textbf{PPL$\downarrow$} & \textbf{COH$\uparrow$} & \textbf{BERT-S$\uparrow$} & \textbf{PPL$\downarrow$} & \textbf{COH$\uparrow$}  \\
\hline
Before & 97.51 & 45.66 & 48.49 & 85.66 & 50.99 & 67.39\\
\hdashline
After & \bf 98.00 & \bf 35.75 & \bf 48.93 & \bf 86.42 & \bf 50.28 & \bf 67.66\\
\hline
\end{tabular}
}
\end{center}
\label{tbl:metric}
\begin{tablenotes}
\item \textit{Note: }The first row displays the performance before refinement, while the second row shows the performance after refinement.
\end{tablenotes}
\end{table}

To examine whether the observed improvements are reflected in both semantic and fluency aspects, we \ignore{use the MuST-C En$\rightarrow$De test set as a representative and }evaluate both the transcription and translation outputs before and after the refinement process. For semantic evaluation, we use BERTScore (BERT-S)\footnote{bert-base-multilingual-cased\_L9\_no-idf\_version=0.3.12(hug\_trans=4.44.1) \\ \_fast-tokenizer}~\cite{zhang_etal_iclr_2020_bert-score} as the prime metric. To assess fluency, we use Perplexity (PPL) and Coherence (COH)~\cite{li_etal_acl_2023_contrastive} as the metrics. Specifically, we calculate perplexity scores with the GPT-2~\cite{gpt2} model and coherence scores with the SimCSE~\cite{gao-etal-emnlp-2021-simcse} model. The results, as presented in Table \ref{tbl:metric}, indicate improvements in both semantic accuracy and fluency. These findings suggest that the refinements enhance not only the fluency of the text but also its underlying semantic meaning.

\subsection{Effect of ASR Transcription Quality}
\label{sec:ana:asr_quality}

The source sentences for \texttt{Refine}$_{Both}$ are derived from noisy ASR outputs. To analyze the impact of ASR transcription quality, we use the context-agnostic fine-tuned LLaMA3-8B model. We generate noisy transcriptions with varying levels of WER using several off-the-shelf Whisper models, ranging from tiny to large~\cite{radford_etal_icml_2023_whisple}. For comparison, we also include gold transcriptions, which represent the ideal case (i.e., Oracle with a 0.00 WER score). Table \ref{tbl:asr_quality} shows the performance across different ASR transcription qualities. As the quality of the ASR transcription improves, both translation and transcription refinement performance steadily improves as well. The model achieves the best results with gold transcriptions, likely due to the reduced noise, which allows for more accurate error detection and correction.

\begin{table}[!t]
\caption{Performance on MuST-C En$\rightarrow$De test set with different ASR transcription qualities. }
\begin{center}
\small
\begin{tabular}{l|rrrr}
\hline
\textbf{Whisper} & \textbf{WER$^\diamond$$\downarrow$} & \textbf{BLEU$\uparrow$} & \textbf{COMET$\uparrow$} & 
\textbf{WER$\downarrow$}\\
\hline
- & - & 29.40* & .7669* & - \\
\hline
Tiny & 13.50 & 31.50 & .8185 & 10.51 \\
Base & 11.22 & 32.11 & .8269 & 8.88 \\
Small & 9.59 & 32.53 & .8311 & 7.68 \\
Medium & 8.95 & 32.87 & .8338 & 7.13 \\
Large & 8.22 & 32.93 & .8360 & 6.63 \\
\hline
Oracle & \bf 0.00 & \bf 34.30 & \bf .8535 & \bf 1.20 \\
\hline
\end{tabular}
\end{center}
\label{tbl:asr_quality}
\begin{tablenotes}
\item \textit{Note: }$\diamond$ and * indicate ASR/ST performance before refinement, respectively.
\end{tablenotes}
\end{table}

\subsection{Effect of Two-Stage Fine-Tuning}
\label{sec:ana:two_stage}

To validate the effectiveness of the two-stage strategy, we compare models fine-tuned with Stage 2 ($S_2$) alone to those fine-tuned with both Stage 1 and Stage 2 ($S_1+S_2$). \ignore{Experiments are conducted with LLaMA3-8B and Mistral-12B for both the context-agnostic and context-aware ($K=3$).} As shown in Table~\ref{tbl:two_stage}, the two-stage fine-tuning strategy consistently outperforms fine-tuning with Stage 2 alone in terms of BLEU scores for both LLaMA3-8B and Mistral-12B. Additionally, the two-stage strategy generally leads to modest improvements in COMET and WER scores.

\begin{table}[!t]
\caption{Performance on MuST-C En$\rightarrow$De test set with single ($S_2$) or two-stage ($S_1+S_2$) fine-tuning strategies.}
\begin{center}
\small
\begin{tabular}{l|l|lll}
\hline
\textbf{Model} & \textbf{Stage} & \textbf{BLEU$\uparrow$} & \textbf{COMET$\uparrow$} & \textbf{WER$\downarrow$} \\
\hline
\multicolumn{5}{c}{\bf Context-Agnostic Fine-Tuning}\\
\hline
\multirow{2}{*}{LLaMA3-8B} & $S_2$ & 31.31 & .8172 & 9.44 \\
 & $S_1 + S_2$ & \bf 32.04 & \bf .8196 & \bf 9.27 \\
\hdashline
\multirow{2}{*}{Mistral-12B} & $S_2$ &  32.40 & \bf .8216 & 9.32 \\
 & $S_1 + S_2$ & \bf 32.75 & \bf .8216 & \bf 9.29 \\
\hline
\multicolumn{5}{c}{\bf Context-Aware Fine-Tuning ($K=3$)}\\
\hline
\multirow{2}{*}{LLaMA3-8B} & $S_2$ & 32.05 & .8213 & 9.06 \\
 & $S_1 + S_2$ & \bf 32.43 & \bf .8240 & \bf 8.95 \\
\hdashline
\multirow{2}{*}{Mistral-12B} & $S_2$ & 33.00 & .8284 & \bf 8.90 \\
 & $S_1 + S_2$ & \bf 33.39 & \bf .8294 & 8.91 \\
\hline
\end{tabular}
\end{center}
\label{tbl:two_stage}
\end{table}

\subsection{Effect of Context Length}
\label{sec:ana:context_len}

In this analysis, we investigate the impact of varying document-level context length on context-aware fine-tuning. We evaluate the performance of the LLaMA3-8B model across different sentence counts, $K \in \{1, 3, 5, 7, 9\}$. As shown in Table \ref{tbl:context_length}, incorporating document-level context results in better performance compared to the absence of it (i.e., $K=1$ for context-agnostic). The model achieves its best results when $K=3$. Notably, increasing the context length beyond $K = 3$ does not result in further performance improvements. Based on these findings, we select $K = 3$ for all context-aware experiments, as it offers the best balance between performance gains and computational efficiency.

\begin{table}[!t]
\caption{Performance on MuST-C En$\rightarrow$De validation set with different context length.}
\begin{center}
\small
\begin{tabular}{l|lll}
\hline
\textbf{$K$} & \textbf{BLEU$\uparrow$} & \textbf{COMET$\uparrow$} & \textbf{WER$\downarrow$} \\
\hline
1 & 30.29 & .8059 & 10.69\\
3 & \bf 31.76 & \bf .8117 & \bf 10.09\\
5 & 31.75 & .8094 & 10.94 \\
7 & 31.30 & .8050 & 10.24 \\
9 & 31.55 & .8065 & 10.84\\
\hline
\end{tabular}
\end{center}
\label{tbl:context_length}
\begin{tablenotes}
\item \textit{Note: }When $K$ is set to 1, the model becomes context-agnostic, as it does not incorporate neighboring sentences. 
\end{tablenotes}
\end{table}

\subsection{Analysis of Context Sensibility}
\label{sec:ana:context_sen}

In Section \ref{sec:aware_fine_tuning}, we use chunk-based decoding (CBD) for context-aware inference. To verify whether context-aware fine-tuning truly takes advantage of contextual information, we follow ~\citet{sun_etal_acl_2022_rethinking} by deliberately introducing incorrect context. Specifically, we begin by shuffling the sentence order within each document and then reassembling it using CBD, which we term \textit{Local Shuffle}, and then, by swapping sentences between documents, which we refer to as \textit{Global Shuffle}. The underlying intuition is that if the model is not sensitive to discourse dependencies, we would expect its performance to remain relatively unaffected by these context shuffling manipulations. To test this hypothesis, we conduct experiments using the LLaMA3-8B model and include the APT metric \cite{miculicich_etal_discomt_2017_validation} to evaluate pronoun translation accuracy. As shown in Table \ref{tbl:context_sensibility}, the results clearly indicate that randomizing context—whether by shuffling sentences within the same document or across different documents—leads to a noticeable decline in performance. Furthermore, using context from different documents (i.e., \textit{Global Shuffle}) results in a more significant performance degradation than shuffling within the same document (i.e., \textit{Local Shuffle}). These findings suggest that the model does indeed effectively  modeling document-level context, highlighting the importance of preserving discourse dependencies in context-aware fine-tuning.

\begin{table}[!t]
\caption{Performance on MuST-C En$\rightarrow$De test set with different document-level context.}
\begin{center}
\small
\begin{tabular}{l|llll}
\hline
\textbf{Shuffle} & \textbf{BLEU$\uparrow$} & \textbf{COMET$\uparrow$} & \textbf{APT$\uparrow$} & \textbf{WER$\downarrow$} \\
\hline
- & \bf 32.43 & \bf .8240 & \bf 68.58 & \bf 8.95 \\
\hline
Local & 31.70 & .8162 & 68.16 & 9.43\\
Global & 31.26 & .8145 & 67.79 & 9.69\\
\hline
\end{tabular}
\end{center}
\label{tbl:context_sensibility}
\end{table}

\subsection{GPT Evaluation}
\label{sec:ana:gpt_eval}

\begin{figure}[t]
\centering
\includegraphics[width=0.40\textwidth, trim={0cm 0cm 0cm 0cm}, clip]{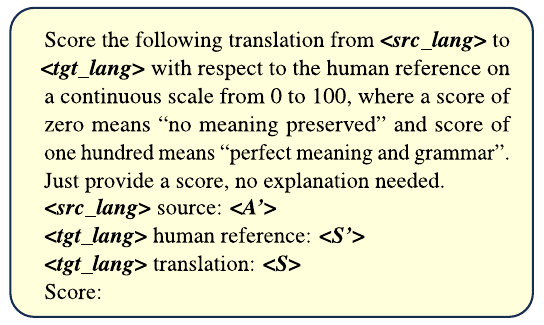}
\caption{Prompt for GPT evaluation.}
\label{fig:gpt_eval_template}
\end{figure}

\begin{table*}[!t]
\caption{Average GPT score for speech translation quality.}
\begin{center}
\small
\begin{tabular}{l|lll|llll}
\hline
\multirow{2}{*}{\textbf{}} & \multicolumn{3}{c|}{\textbf{MuST-C}} &  \multicolumn{4}{c}{\textbf{CoVoST 2}} \\
\bf Refinement & {\textbf{En$\rightarrow$De}} & {\textbf{En$\rightarrow$Fr}} & {\textbf{En$\rightarrow$Es}} & {\textbf{En$\rightarrow$De}} & {\textbf{En$\rightarrow$Ca}} & {\textbf{En$\rightarrow$Ar}} & {\textbf{En$\rightarrow$Tr}} \\
\hline
Before & 73.58 & 78.89 & 76.82 & 69.00 & 71.83 & 73.38 & 70.15 \\
\hdashline
After & \bf 78.92 & \bf 82.83 & \bf 80.92 & \bf 79.24 & \bf 80.88 & \bf 77.80 & \bf 76.53 \\
\hline
\end{tabular}
\end{center}
\label{tbl:quality_assessment}
\begin{tablenotes}
\item \textit{Note: }The first row displays the performance before refinement, while the second row shows the performance after refinement.
\end{tablenotes}
\end{table*}

Based on the \textit{Direct Assessment} prompt from ~\citet{kocmi_federmann_EAMT_2023_large}, as shown in Figure \ref{fig:gpt_eval_template}, we use GPT-4o to evaluate speech translation quality on a 0-100 scale for both the initial and refined translations. Our in-house dataset consists of 200 randomly selected samples from \texttt{Refine}$_{Both}$ task of Mistral-12B under sentence-level fine-tuning setting. As shown in Table \ref{tbl:quality_assessment}, for the seven language pairs in the MuST-C and CoVoST 2 datasets, GPT-4o consistently assign higher evaluation scores to the refined translations compared to the initial ones. This improvement is particularly notable for  En$\rightarrow$De and En$\rightarrow$Ca translation refinement, with score increases of 10.24 and 9.05, respectively. These results align with other metrics (e.g. BLEU and COMET), which similarly indicate that the refined translations are superior to the initial ones. We will provide specific cases in Section \ref{sec:ana:case_study} for further illustration.

\subsection{Refining Speech Translation from Different System}
\label{sec:ana:result_other_system}

In Section~\ref{sec:experiment}, we refine speech translation (ST) using the MuST-C dataset with our our implementation of CRESS~\cite{fang_feng_acl_2023_cress}, a state-of-the-art end-to-end ST system. To further demonstrate the robustness of our approach across different translation qualities, we also refine ST using ConST~\cite{ye_etal_naacl_2022_const} in this section. Specifically, we first generate ASR and ST data for refinement by performing inference on the MuST-C \texttt{tst-COMMON} test set using the open-source ConST model\footnote{\url{https://huggingface.co/ReneeYe/ConST_en2x_models}}. As show in Table \ref{tbl:const_result}, for the \texttt{Refine}$_{Both}$ task without document-level context, Mistral-12B achieves significant improvements across several metrics. Specifically, it shows absolute increases of 3.67, 4.32, and 3.54 in BLEU scores, 0.0517, 0.0474, and 0.0416 in COMET scores, and 1.64, 1.29, and 1.53 in WER scores for En$\rightarrow$De, En$\rightarrow$Fr, and En$\rightarrow$Es translation refinements, respectively. When document-level context is incorporated, Mistral-12B further achieves an additional improvement of 0.49, 0.49, and 0.15 in BLEU scores, and 0.1040, 0.0075, and 0.0080 in COMET scores for En$\rightarrow$De, En$\rightarrow$Fr, and En$\rightarrow$Es translation refinement, respectively. As in Section \ref{sec:result_mustc}, we observe that joint transcription (i.e., \texttt{Refine}$_{Both}$) leads to higher BLEU scores and comparable COMET scores, along with an advantage of improving transcription compared to refining only ST (i.e., \texttt{Refine}$_{ST}$). Overall, these results confirm that our model is capable of refining ST across a variety of translation qualities\ignore{, even in the absence of specialized fine-tuning}.

\begin{table*}[!t]
\caption{Experimental results of refining ST from ConST.}
\begin{center}
\resizebox{\linewidth}{!}{
\begin{tabular}{l|l|lll|lll|lll}
\hline
\multirow{2}{*}{\textbf{Task}} & \multirow{2}{*}{\textbf{Model}} & \multicolumn{3}{c|}{\textbf{En$\rightarrow$De}} & \multicolumn{3}{c|}{\textbf{En$\rightarrow$Fr}} & \multicolumn{3}{c}{\textbf{En$\rightarrow$Es}} \\
& & \textbf{BLEU$\uparrow$} & \textbf{COMET$\uparrow$} & \textbf{WER$\downarrow$} & \textbf{BLEU$\uparrow$} & \textbf{COMET$\uparrow$} & \textbf{WER$\downarrow$} & \textbf{BLEU$\uparrow$} & \textbf{COMET$\uparrow$} & \textbf{WER$\downarrow$} \\
\hline
 & ConST & 28.10 & .7624 & 12.31 & 38.15 & .7701 & 11.49 & 31.91 & .7748 & 11.38\\
\hline
\multicolumn{11}{c}{\bf Context-Agnostic Fine-Tuning}\\
\hline
\multirow{2}{*}{\texttt{Refine}$_{Both}$} & LLaMA3-8B & 30.58 & .8083 & 10.82 & 41.30 & .8125	& 10.37 & 34.95	& .8109 & 10.16 \\
& Mistral-12B & \bf 31.77 & \bf .8141 & \bf 10.67 & \bf 42.47 & \bf .8175 & \bf 10.20 & \bf 35.45 & \bf .8164 & \bf 9.85 \\
\hdashline
\multirow{2}{*}{\texttt{Refine}$_{ST}$} & LLaMA3-8B & 29.23 & .8081 & - & 38.74 & .8114 & - & 33.56 & .8137 & - \\
& Mistral-12B & \bf 30.14 & \bf .8154 & - & \bf 41.51 & \bf .8196 & - & \bf 34.91 & \bf .8184 & - \\
\hline
\multicolumn{11}{c}{\bf Context-Aware Fine-Tuning ($K=3$)}\\
\hline
\multirow{2}{*}{\texttt{Refine}$_{Both}$} & LLaMA3-8B & 31.17 & .8165 & 10.88 & 42.24 & .8196 & 10.16 & 34.85 & .8188 & \bf 9.91 \\
& Mistral-12B & \bf 32.26 & \bf .8245 & \bf 10.74 & \bf 42.96	& \bf .8250	& \bf 10.10 & \bf 35.60 & \bf .8244 & 10.15 \\
\hdashline
\multirow{2}{*}{\texttt{Refine}$_{ST}$} & LLaMA3-8B & 30.65 & .8153 & - & 41.03 & .8190 & - & 34.16 & .8196 & - \\
& Mistral-12B & \bf 31.06 & \bf .8234 & - & \bf 42.86 & \bf .8234 & - & \bf 34.55 & \bf .8232 & - \\
\hline
\end{tabular}
}
\end{center}
\label{tbl:const_result}
\end{table*}

\subsection{Case Analysis}
\label{sec:ana:case_study}

In this section, we provide a detailed comparison of specific cases before and after the refinement process. We examine three examples from the MuST-C En$\rightarrow$De, using CRESS (Ours) and Mistral-12B in the \texttt{Refine}$_{Both}$ task, as shown in Table \ref{tbl:case_study}. These cases demonstrate how the refinement process improves both automatic speech recognition (ASR) and translation outputs.
\begin{itemize}
\item \textbf{Case 1}: In this example, the word ``Mai'' is incorrectly transcribed as ``May''in the automatic transcription. By referring to the word ``Mai'' from the automatic translation, the refinement process updates the transcription to the correct word ``Mai''. Similarly, the word ``angerodneten'' in the automatic translation is refined to ``arrangierten'' in the refined translation. This correction likely occurs by cross-referencing the correct English word ``arranged'' from either the automatic transcription or the refined translation.
\item \textbf{Case 2}: In this example, the word ``Intersexuelle'' in the automatic translation helps to correct the erroneous transcription of ``intersects'' to the accurate term ``intersex'' in the refined transcription. Additionally, the phrase ``dieforschung von Geschlechtsunterschieden'' in the automatic translation contains two issues: the misspelled word ``dieforschung'' and the awkward use of ``von''. By leveraging the automatic translation and the phrase ``sex-difference research'' from both the automatic and refined transcriptions, the word ``dieforschung'' is corrected to ``die Forschung'', and ``von'' is replaced with ``über'', resulting in a more natural and accurate expression.
\item \textbf{Case 3}: In this example, the phrase ``unsere ist'' from the automatic translation is used to correct the transcription ``as ours'' to ``is ours'' in the refined transcription. Additionally, the unnecessary word ``in'' in the phrase ``this interactive world'' from the automatic translation is removed during refinement, leading to a cleaner and more precise translation.
\end{itemize}

These cases highlight how the refinement process improves translation accuracy by addressing errors in both transcription and translation outputs. By leveraging information from both the automatic transcription and translation, our approach significantly enhances the overall quality of speech translation.

\begin{table*}[!ht]
\caption{Examples of correcting ASR and ST errors from the MuST-C En$\rightarrow$De test set.}
\centering
\small
\begin{tabular}{lp{12cm}}
\hline
\multicolumn{2}{c}{\textbf{CASE 1}}  \\
\hline
Auto Transcription & My mother, \textbf{May}, was 18 when her father died, already in an \uline{arranged} marriage already with two small girls.                   \\
Refined Transcription &  My mother, \uline{Mai}, was 18 when her father died, already in an \uline{arranged} marriage, already with two small girls.                   \\
Gold Transcription & My mother, \uline{Mai}, was 18 when her father died — already in an \uline{arranged} marriage, already with two small girls. \\
\hline
Auto Translation &  Meine Mutter, \uline{Mai}, war 18, als ihr Vater starb, bereits in einer \textbf{angeordneten} Ehe, bereits mit zwei kleinen Mädchen.                   \\
Refined Translation & Meine Mutter, \uline{Mai}, war 18, als ihr Vater starb, bereits in einer \uline{arrangierten} Ehe, bereits mit zwei kleinen Mädchen.                    \\
Gold Translation & Meine Mutter \uline{Mai} war 18, als ihr Vater starb – schon in einer \uline{arrangierten} Ehe, schon mit zwei kleinen Mädchen. \\
\hline
\multicolumn{2}{c}{\textbf{CASE 2}}              \\
\hline
Auto Transcription &  For years, because I've been interested in \textbf{intersects}, I've also been interested in \uline{sex-difference research}.                   \\
Refined Transcription &  For years, because I've been interested in \uline{intersex}, I've also been interested in \uline{sex-difference research}.                   \\
Gold Transcription &   For years, because I've been interested in \uline{intersex}, I've also been interested in \uline{sex-difference research}.                  \\
\hline
Auto Translation & Jahrelang, weil ich mich für \uline{Intersexuelle} interessiert habe, habe ich mich auch für \textbf{dieforschung von Geschlechtsunterschieden} interessiert.                     \\
Refined Translation &  Jahrelang, weil ich mich für \uline{Intersexuelle} interessiert habe, habe ich mich auch für \uline{die Forschung über Geschlechtsunterschiede} interessiert.                   \\
Gold Translation &  Seit Jahren schon, weil ich an \uline{Intersexualität} interessiert war, habe ich mich auch für \uline{Forschung im Bereich der Geschlechterdifferenz} interessiert.               \\
\hline
\multicolumn{2}{c}{\textbf{CASE 3}}              \\
\hline
Auto Transcription &  We want to encourage a world of creators, of inventors, of contributors, because this world that we live in, \uline{this interactive world}, \textbf{as ours}.           \\
Refined Transcription &  We want to encourage a world of creators, of inventors, of contributors, because this world that we live in, \uline{this interactive world}, \uline{is ours}.                   \\
Gold Transcription &    We want to encourage a world of creators, of inventors, of contributors, because this world that we live in, \uline{this interactive world}, \uline{is ours}.                 \\
\hline
Auto Translation &   Wir wollen eine Welt der Schöpfer, der Erfinder, der Mitwirkenden fördern, weil diese Welt, in der wir leben, \textbf{in dieser interaktiven Welt}, \uline{unsere ist}.                  \\
Refined Translation &  Wir wollen eine Welt der Schöpfer, der Erfinder, der Mitwirkenden fördern, weil diese Welt, in der wir leben, \uline{diese interaktive Welt}, \uline{unsere ist}.                   \\
Gold Translation &  Wir wollen die Welt der Erschaffer, der Erfinder, der Mitwirkenden ermutigen, denn diese Welt, in der wir leben, \uline{diese interaktive Welt}, \uline{gehört uns}.                   \\
\hline
\end{tabular}
\label{tbl:case_study}
\begin{tablenotes}
\item \textit{Note: }\textbf{Bold} text indicates incorrect words or phrases and \uline{underlined} text indicates accurate words or phrases.
\end{tablenotes}
\end{table*}

\section{Conclusion}

This paper explores how large language models (LLMs) can improve speech translation by simultaneously refining both transcription and translation using in-context learning and task-specific fine-tuning techniques. We begin by designing prompts tailored for translation refinement and evaluating their effectiveness within an in-context learning framework. Next, we fine-tune the LLaMA3-8B and Mistral-12B models, both with and without document-level context, to further enhance the refinement process. Experimental results on both the MuST-C and CoVoST 2 datasets demonstrate that jointly refining transcription and translation results in significant improvements in speech translation quality. These findings highlight the advantages of combining both translation and transcription refinement to achieve superior overall performance. In future work, we will investigate the use of speech inputs for refining speech translation, which could lead to further improvements in translation quality.

\ignore{\section{Limitations}
Due to resource limitations, our experiments focus exclusively on English-to-\textit{X} language pairs. While jointly refining ASR and ST outputs can correct errors in the text, it becomes challenging for LLM to draw insights from ASR and ST outputs when they are heavily noisy. In such cases, the lack of gold information (e.g., speech) makes it difficult to perform effective translation refinement. \revise{Similar to other methods involving translation refinement or post-editing, our approach introduces latency while achieving significant improvements in speech translation quality after the refinement process.} }

{\appendices

\ignore{
\section{Translation Performance}
\label{apdx:translation_performance}

\begin{table}[!t]
\caption{Translation Performance on MuST-C En$\rightarrow$De test set.}
\begin{center}
\small
\begin{tabular}{lll}
\toprule
\textbf{Task} & \textbf{BLEU$\uparrow$} & \textbf{COMET$\uparrow$} \\
\midrule
$\left(A, S\right) \rightarrow \left(A', S'\right)$ & 32.04 & .8196 \\
\midrule
$\left(A'\right) \rightarrow \left(S'\right)$ & \bf 35.31 & \bf .8606 \\
$\left(A\right) \rightarrow \left(S'\right)$ & 30.27 & .8079  \\
\bottomrule
\end{tabular}
\end{center}
\label{tbl:translation}
\end{table}

To evaluate the performance of our proposed approach relative to text-to-text translation models, we conducted a series of fine-tuning experiments utilizing Llama3-8B on the MuST-C En$\rightarrow$De dataset, focusing on distinct tasks. These results are presented in Table \ref{tbl:translation}.  Specifically, $\left(A, S\right) \rightarrow \left(A', S'\right)$ represents the Refine$_{Both}$ task detailed in the section \ref{sec:task}, $\left(A'\right) \rightarrow \left(S'\right)$ denotes the text-to-text translation task, and $\left(A\right) \rightarrow \left(S'\right)$ mirrors $\left(A'\right) \rightarrow \left(S'\right)$ but differs in the transcription used during decoding (e.g. unrefined transcription vs gold transcription). As indicated in the table \ref{tbl:translation}, our refinement model exhibits performance that falls short of the text-to-text translation model employing gold transcription, yet surpasses the text-to-text translation model utilizing unrefined transcription. This outcome is attributable to two primary factors: firstly, the source input for our refinement model inherently contains more noise compared to the gold transcription used in the text-to-text translation baseline; secondly, unlike the text-to-text translation model that relies solely on unrefined transcription, our refinement model leverages both the unrefined transcription and its corresponding translation to simultaneously correct the transcription and translation, enhancing the quality of refined translation.
}

\section{Detailed Performance on CoVoST 2}
\label{apdx:result_covost2}

Since using the GPT-3.5-turbo API can be expensive, we randomly select 500 samples from each translation task in the CoVoST 2 test set to create our in-house test set. Table~\ref{tbl:covost2_gpt_result} shows the performance on this in-house test set. For \texttt{Refine$_{Both}$} and \texttt{Refine$_{ST}$}, in-context learning enhances the performance of ST refinement, and in few-shot settings, it also leads to noticeable improvements in transcription refinement. However, \texttt{Paraphrase$_{ST}$} lags behind both \texttt{Refine$_{Both}$} and \texttt{Refine$_{ST}$} in overall performance, even falling below the baseline.

\begin{table*}[!t]
\caption{
Performance of GPT-3.5 with Different In-Context Learning Strategies on the in-house CoVoST 2 test set.}
\begin{center}
\resizebox{\linewidth}{!}{
\begin{tabular}{l|l|lll|lll|lll|lll}
\hline
\multirow{2}{*}{\textbf{Task}} & \multirow{2}{*}{\textbf{Model}} & \multicolumn{3}{c|}{\textbf{En$\rightarrow$De}} & \multicolumn{3}{c|}{\textbf{En$\rightarrow$Ca}} & \multicolumn{3}{c|}{\textbf{En$\rightarrow$Ar}} & \multicolumn{3}{c}{\textbf{En$\rightarrow$Tr}}\\
& & \textbf{BLEU$\uparrow$} & \textbf{COMET$\uparrow$} & \textbf{WER$\downarrow$} & \textbf{BLEU$\uparrow$} & \textbf{COMET$\uparrow$} & \textbf{WER$\downarrow$} & \textbf{BLEU$\uparrow$} & \textbf{COMET$\uparrow$} & \textbf{WER$\downarrow$} & \textbf{BLEU$\uparrow$} & \textbf{COMET$\uparrow$} & \textbf{WER$\downarrow$} \\
\hline
- & SpeechLM-P / Whisper & 28.73 & .7502 & 11.25 & 34.73 & .7618 & 10.36 & 21.61 & .7924 & 10.36 & 19.06 & .8013  & 10.36  \\
\hline
\multicolumn{14}{c}{In-Context Learning}\\
\hline
\multirow{7}{*}{\texttt{Refine}$_{Both}$} & GPT-3.5 (0) & 36.38 & .8446 & 12.42 & 40.23 & .8275 & 12.12 & 23.04 & \bf .8385 & 11.36 & 21.56  & \bf .8535 & 12.03 \\
& GPT-3.5 (R1) & 36.14 & \bf .8467 & 11.14 & 40.27 & .8267 & 10.29 & 23.70 & .8347 & 10.27	& 22.41 & .8487 & 10.27 \\
& GPT-3.5 (R3) & 36.01 & .8451 & 11.30 & 40.60 & .8270 & 10.07 & \bf 24.95 & .8381 & 9.96 & 21.49 & .8466 & 10.12 \\ 
& GPT-3.5 (R5) & \bf 36.56 & .8396 & 10.94 & 40.64 & .8262 & 10.00 & 24.42 & .8344	& \bf 9.94 & 21.91 & .8473 & 9.92 \\
& GPT-3.5 (T1) & 35.62 & .8433 & 11.05 & \bf 41.40 & \bf .8293 & 10.38 & 23.00 & .8337 & 10.47 & \bf 22.58 & .8471 & 9.89 \\
& GPT-3.5 (T3) & 36.21 & .8416 & \bf 10.65 & 40.29 & .8247 & 10.07 & 23.65 & .8331 & 10.12 & 22.25 & .8492 & 9.85 \\
& GPT-3.5 (T5) & 35.26 & .8403 & 10.74 & 40.90 & .8228 & \bf 9.74 & 24.13 & .8305 & 9.96 & 22.04 & .8463 & \bf 9.72 \\
\hdashline
\multirow{7}{*}{\texttt{Refine}$_{ST}$} & GPT-3.5 (0) & 34.60 & .8417 & - & 39.54 & .8242 & - & 21.98 & .8329 & - & 21.12 & \bf .8461 & - \\
& GPT-3.5 (R1) & 36.23 & .8438 & - & 39.45 & .8229 & - & 23.23 & .8337 & - & 22.13 & .8435 & - \\
& GPT-3.5 (R3) & 35.05 & .8441 & - & 40.12 & .8253 & - & 23.28 & \bf .8352 & - & 21.92 & .8439 & - \\
& GPT-3.5 (R5) & \bf 36.87 & .8435 & - & \bf 40.65 & \bf .8261 & - & \bf 24.06 & .8338 & - & \bf 22.47 & .8445 & - \\
& GPT-3.5 (T1) & 35.55 & \bf .8451 & - & 39.87 & .8243 & - & 22.91 & .8326 & - & 22.06 & .8448 & - \\
& GPT-3.5 (T3) & 36.04 & .8437 & - & 40.28 & .8248 & - & 23.42 & .8322 & - & 21.77 & .8452 & - \\
& GPT-3.5 (T5) & 35.70 & .8387 & - & 39.88 & .8252 & - & 23.23 & .8322 & - & 21.94 & .8447 & - \\
\hdashline
\multirow{4}{*}{\texttt{Paraphrase}$_{ST}$} & GPT-3.5 (0) & 16.68 & .7728 & - & 18.04 & .7042 & - & 12.27 & .7755 & - & 9.86 & .7902 & - \\
& GPT-3.5 (T1) & 26.14 & \bf .7746 & - & 29.72 & .7631 & - & 17.69 & .7876 & - & 15.14 & .8016 & - \\
& GPT-3.5 (T3) & 28.81 & .7677 & - & 33.69 & .7684 & - & 19.31 & .7898 & - & 17.02 & .8061 & - \\
& GPT-3.5 (T5) & \bf 29.73 & .7649 & - & \bf 34.56 & \bf .7690 & - & \bf 20.12 & \bf .7920 & - & \bf 17.55 & \bf .8080 & - \\
\hline
\end{tabular}
}
\end{center}
\label{tbl:covost2_gpt_result}
\end{table*}

In addition to the closed-source GPT-3.5-turbo, we use the \texttt{Refine}$_{Both}$ task as a representative task to conduct experiments with the open-source Mistral-12B model on the in-house CoVoST 2 test set. The results, presented in Table \ref{tbl:mistral_icl}, show that, compared to the baseline system, different prompt selection strategies lead to improvements in both BLEU and COMET scores across all four language pairs. This demonstrates that Mistral can effectively refine translations without explicit fine-tuning, relying solely on its in-context learning capabilities.

\ignore{
\section{In-context learning for Refine$_{Both}$}
Based on the results presented in Tables \ref{tbl:mustc_result} and \ref{tbl:covost2_result}, Mistral-12B outperforms the closed-source GPT-3.5-turbo in fine-tuned scenarios and achieves the best overall performance on the Refine$_{Both}$ task. To further assess in-context learning capabilities of the open-source model, we conduct experiments leveraging Mistral-12B model on the Refine$_{Both}$ task on the in-house CoVoST 2 test set. The results, as shown in Table \ref{tbl:mistral_icl}, demonstrate the following: 
1) When compared to the baseline system, various prompt selection strategies result in improvements in both BLEU and COMET scores across all four language pairs, indicating that Mistral can effectively refine translations even without explicit fine-tuning, relying solely on its in-context learning abilities. 2) In contrast to the baseline system, including GPT-3.5-turbo, a deterioration in WER scores is observed across all language pairs, which may suggest that, in the absence of fine-tuning, the model struggles to utilize the translation text for correcting transcription errors. 3) Overall, Mistral’s refinement performance remains inferior to that of zero-shot GPT-3.5-turbo, highlighting an ongoing gap between open-source and closed-source models, particularly in terms of in-context learning capabilities.
}

\begin{table*}[!t]
\caption{
Performance of Mistral with Different In-Context Learning Strategies for Refine$_{Both}$ on the in-house CoVoST 2 test set.}
\begin{center}
\resizebox{\linewidth}{!}{
\begin{tabular}{l|lll|lll|lll|lll}
\hline
\multirow{2}{*}{\textbf{Model}} & \multicolumn{3}{c|}{\textbf{En$\rightarrow$De}} & \multicolumn{3}{c|}{\textbf{En$\rightarrow$Ca}} & \multicolumn{3}{c|}{\textbf{En$\rightarrow$Ar}} & \multicolumn{3}{c}{\textbf{En$\rightarrow$Tr}}\\
& \textbf{BLEU$\uparrow$} & \textbf{COMET$\uparrow$} & \textbf{WER$\downarrow$} & \textbf{BLEU$\uparrow$} & \textbf{COMET$\uparrow$} & \textbf{WER$\downarrow$} & \textbf{BLEU$\uparrow$} & \textbf{COMET$\uparrow$} & \textbf{WER$\downarrow$} & \textbf{BLEU$\uparrow$} & \textbf{COMET$\uparrow$} & \textbf{WER$\downarrow$} \\
\hline
SpeechLM-P / Whisper & 28.73 & .7502 & \bf 11.25 & 34.73 & .7618 & \bf 10.36 & 21.61 & .7924 & \bf 10.36 & 19.06 & .8013  & \bf 10.36  \\
GPT-3.5 (0) & \bf 36.38 & \bf .8446 & 12.42 & \bf 40.23 & \bf .8275 & 12.12 & 23.04 & \bf .8385 & 11.36 & \bf 21.56  & \bf .8535 & 12.03 \\
\hdashline
Mistral-12B (0) & 33.24	& .8349	& 16.03	& 38.11	& .8082 & 14.87 & 23.32 & .8202 & 15.32 & 19.51 & .8244 & 15.36  \\
Mistral-12B (R1) & 34.19 & .8406 & 15.62 & 38.12 & .8096 & 13.47 & 23.56 & .8192 & 15.23 & 19.26 & .8288 & 13.78  \\
Mistral-12B (R3) & 34.06 & .8387 & 14.99 & 38.72 & .8130 & 13.58 & 23.54 & .8190 & 13.87 & 19.21 & .8297 & 13.83  \\
Mistral-12B (R5) & 34.11 & .8355 & 13.59 & 38.55 & .8156 & 13.14 & 23.36 & .8212 & 13.05 & 19.98 & .8357 & 13.32  \\
Mistral-12B (T1) & 34.11 & .8378 & 14.52 & 38.71 & .8116 & 13.61 & 23.06 & .8196 & 14.34 & 20.31 & .8325 & 14.50  \\
Mistral-12B (T3) & 34.41 & .8394 & 13.12 & 38.86 & .8078 & 12.41 & 23.45 & .8239 & 12.16 & 20.96 & .8330 & 12.23  \\
Mistral-12B (T5) & 35.05 & .8370 & 12.90 & 39.02	& .8114 & 11.89 & \bf 24.57 & .8254 & 12.07 & 20.99 & .8313 & 12.21 \\
\hline
\end{tabular}
}
\end{center}
\label{tbl:mistral_icl}
\end{table*}

\section{Prompt Examples}
\label{apdx:prompt_example}

Table \ref{tbl:prompt_example_refine_both} to Table \ref{tbl:prompt_example_paraphrase_st} illustrate prompts used in tasks of \texttt{Refine}$_{Both}$, \texttt{Refine}$_{ST}$, and \texttt{Paraphrase}$_{ST}$, respectively.

\begin{table*}[!h]
\caption{A prompt example of \texttt{Refine$_{Both}$}, with gray areas indicating the in-context learning examples.}
\begin{center}
\resizebox{\textwidth}{!}{
\begin{tabular}{p{20cm}}
\hline
Given the \textit{English} transcription and \textit{German} translation, both derived from speech and potentially containing errors, please provide the refined transcription and translation without any explanation. Present the results in two lines, starting with ``Refined Transcription:'' and ``Refined Translation:'', respectively. 
\\
\begin{tabular}[c]{@{}l@{}}
\colorbox[gray]{0.8}{Let me give you \textit{2} examples.}\\
\colorbox[gray]{0.8}{\#\# 1}\\
\colorbox[gray]{0.8}{Transcription: \textit{Now, there's a lot going on in this movie, so let me break this down and show you what's going on.
}}\\
\colorbox[gray]{0.8}{Translation: \textit{Es gibt viel in diesem Film, also lassen Sie mich das zerlegen und Ihnen zeigen, was los ist.
}}\\
\colorbox[gray]{0.8}{Refined Transcription: \textit{Now, there's a lot going on in this movie, so let me break this down and show you what's going on.
}}\\ \colorbox[gray]{0.8}{Refined Translation: \textit{Also, in diesem Film passiert sehr viel, also lassen Sie mich das analysieren und Ihnen zeigen was passiert.
}}\\
\colorbox[gray]{0.8}{\#\# 2}\\
\colorbox[gray]{0.8}{Transcription: \textit{So I'm going to show you a movie where you're going to see that kind of dynamic.}}\\
\colorbox[gray]{0.8}{Translation: \textit{Ich werde Ihnen einen Film zeigen, in dem Sie diese Art von Dynamik sehen werden.}}\\
\colorbox[gray]{0.8}{Refined Transcription: \textit{So I'm going to show you a movie where you're going to see that kind of dynamic.}}\\ 
\colorbox[gray]{0.8}{Refined Translation: \textit{Ich werde nun einen Film zeigen, in dem Sie diesen Vorgang sehen.
}}\\
\colorbox[gray]{0.8}{Now consider the following transcription and translation, please provide the refined transcription and translation following above} \\ \colorbox[gray]{0.8}{output format.}\\
Transcription: \textit{You're going to see the whole thing take place in this movie.} \\
Translation: \textit{Sie werden sehen, wie das Ganze in diesem Film passiert.} \\
\end{tabular} \\
\hline

\end{tabular}
}
\label{tbl:prompt_example_refine_both}
\end{center}
\end{table*}

\begin{table*}[!h]
\caption{A prompt example of \texttt{Refine$_{ST}$}, with gray areas indicating the in-context learning examples.}
\begin{center}
\resizebox{\textwidth}{!}{
\begin{tabular}{p{20cm}}
\hline
Given the \textit{English} transcription and \textit{German} translation, both derived from speech and potentially containing errors, please provide the refined translation without any explanation. Present the result in one line, starting with ``Refined Translation:''. 
\\
\begin{tabular}[c]{@{}l@{}}
\colorbox[gray]{0.8}{Let me give you \textit{2} examples.}\\
\colorbox[gray]{0.8}{\#\# 1}\\
\colorbox[gray]{0.8}{Transcription: \textit{Now, there's a lot going on in this movie, so let me break this down and show you what's going on.
}}\\
\colorbox[gray]{0.8}{Translation: \textit{Es gibt viel in diesem Film, also lassen Sie mich das zerlegen und Ihnen zeigen, was los ist.
}}\\
\colorbox[gray]{0.8}{Refined Translation: \textit{Also, in diesem Film passiert sehr viel, also lassen Sie mich das analysieren und Ihnen zeigen was passiert.
}}\\
\colorbox[gray]{0.8}{\#\# 2}\\
\colorbox[gray]{0.8}{Transcription: \textit{So I'm going to show you a movie where you're going to see that kind of dynamic.}}\\
\colorbox[gray]{0.8}{Translation: \textit{Ich werde Ihnen einen Film zeigen, in dem Sie diese Art von Dynamik sehen werden.}}\\
\colorbox[gray]{0.8}{Refined Translation: \textit{Ich werde nun einen Film zeigen, in dem Sie diesen Vorgang sehen.
}}\\
\colorbox[gray]{0.8}{Now consider the following transcription and translation, please provide the refined translation following above output format.}\\
Transcription: \textit{You're going to see the whole thing take place in this movie.} \\
Translation: \textit{Sie werden sehen, wie das Ganze in diesem Film passiert.} \\
\end{tabular} \\
\hline

\end{tabular}
}
\label{tbl:prompt_example_refine_st}
\end{center}
\end{table*}

\begin{table*}[!h]
\caption{A prompt example of \texttt{Paraphrase$_{ST}$}, with gray areas indicating the in-context learning examples.}
\begin{center}
\resizebox{\textwidth}{!}{
\begin{tabular}{p{20cm}}
\hline
Please give me a paraphrase in \textit{German} without any explanation. Present the result in one line, starting with ``Paraphrase:''. \\
\begin{tabular}[c]{@{}l@{}}
\colorbox[gray]{0.8}{Let me give you \textit{2} examples.}\\
\colorbox[gray]{0.8}{\#\# 1}\\
\colorbox[gray]{0.8}{Sentence: \textit{Es gibt viel in diesem Film, also lassen Sie mich das zerlegen und Ihnen zeigen, was los ist.
}}\\
\colorbox[gray]{0.8}{Paraphrase: \textit{Also, in diesem Film passiert sehr viel, also lassen Sie mich das analysieren und Ihnen zeigen was passiert.
}}\\
\colorbox[gray]{0.8}{\#\# 2}\\
\colorbox[gray]{0.8}{Sentence: \textit{Ich werde Ihnen einen Film zeigen, in dem Sie diese Art von Dynamik sehen werden.}}\\
\colorbox[gray]{0.8}{Paraphrase: \textit{Ich werde nun einen Film zeigen, in dem Sie diesen Vorgang sehen.
}}\\
\colorbox[gray]{0.8}{Now consider the following sentence, please provide the \textit{German} paraphrase following above output format.}\\
Sentence: \textit{You're going to see the whole thing take place in this movie.} \\
\end{tabular} \\
\hline
\end{tabular}
}
\label{tbl:prompt_example_paraphrase_st}
\end{center}
\end{table*}

}

 
%
 
\bibliography{taslp}
\bibliographystyle{IEEEtranN}

\vfill

\end{document}